\documentclass{article}

\usepackage[preprint,nonatbib]{neurips_2023}
\usepackage{CJKutf8}

\usepackage[utf8]{inputenc} 
\usepackage[T1]{fontenc}    

\usepackage{times}
\usepackage{latexsym}

\usepackage{microtype}
\usepackage{hyperref}
\usepackage{graphicx}
\usepackage{xspace}
\usepackage{paralist}
\usepackage{wrapfig}
\usepackage{booktabs}
\usepackage{multirow}
\usepackage{amsmath}
\usepackage{amsfonts}
\usepackage{amssymb}
\usepackage{pifont}
\usepackage{mathrsfs}
\usepackage{algorithm,algpseudocode}
\usepackage{mdwlist}
\usepackage{enumitem}
\usepackage{xcolor}
\usepackage{dsfont}
\usepackage{amsthm}
\usepackage{bm}
\usepackage{float}
\usepackage{caption}
\usepackage{subcaption}
\usepackage{tcolorbox}

\newcommand{\method}{LLM-ST\xspace}

\title{Speech Translation with Large Language Models: An Industrial Practice}

\author{%
  Zhichao Huang, Rong Ye, Tom Ko, Qianqian Dong, \\
  \textbf{Shanbo Cheng, Mingxuan Wang, Hang Li}\\
  \\
  ByteDance\\
  \texttt{\{zhichao.huang, yerong\}@bytedance.com} \\
}

\begin{document}

\maketitle

\begin{abstract}
Given the great success of large language models~(LLMs) across various tasks, in this paper, we introduce \method, a novel and effective speech translation model constructed upon a pre-trained LLM. 
By integrating the large language model~(LLM) with a speech encoder and employing multi-task instruction tuning, \method can produce accurate timestamped transcriptions and translations, even from long audio inputs. 
Furthermore, our findings indicate that the implementation of Chain-of-Thought (CoT) prompting can yield advantages in the context of LLM-ST.
Through rigorous experimentation on English and Chinese datasets, we showcase the exceptional performance of LLM-ST, establishing a new benchmark in the field of speech translation.
Demo: \url{https://speechtranslation.github.io/llm-st/}
\end{abstract}

\section{Introduction}
Developing a speech translation system, particularly for lengthy speeches, typically involves a complex pipeline. Take, for instance, the task of generating video subtitles, where the conventional procedure typically encompasses the following steps: 
\begin{inparaenum}[\it 1)]
    \item segmenting the speech, 
    \item annotating timestamps, 
    \item transcribing via an automatic speech recognition~(ASR) model, 
    \item inverse text normalization~(ITN) for readability, 
    \item jointly translating the transcription, and finally,
    \item aligning the translated text with the respective timestamps.
\end{inparaenum}
While these tasks share functional similarities, the lengthy pipeline make it preferable to address them collectively with a unified model. To achieve this objective, leveraging a large language model appears to be a promising choice, given its exceptional ability to accomplish diverse tasks.

Recently, the development of large language models (LLMs)~\cite{ouyang2022training,bai2022constitutional,bubeck2023sparks,anil2023palm,touvron2023llama} has surpassed a series of NLP tasks in a single stride.
Specifically, it significantly raised the bar for machine translation (MT), particularly in avoiding literal translations~\cite{raunak2023gpts}, resulting in translations that are more contextually appropriate and readable~\cite{zhang2023prompting,kocmi2023large,wang2023document,raunak2023gpts}.
Concurrently, cross-modality research between speech and text has emerged, with some work focusing on integrating LLMs in the realm of speech.
One branch of the approach is to discretize speech and then extensively pre-train discrete speech tokens using the framework of language models, like SpeechGPT~\cite{zhang2023speechgpt}, PolyVoice~\cite{dong2023polyvoice} and VioLA~\cite{wang2023viola}.
While conceptually appealing, they often encounter tangible performance challenges in downstream tasks such as ASR, ST, and spoken language question-answering.
Another branch introduces continuous speech representations directly to LLMs, enhancing the models with a cohesive blend of ASR and ST capabilities~\cite{fathullah2023prompting, tsunoo2023decoder}.

In this paper, we propose \method.
In terms of the structure, \method is based on a pre-trained language model, giving it superior meaning understanding and translation capabilities. 
We employ a speech encoder to provide continuous speech representations to the LLM, ensuring no loss of information through speech discretization and maintaining the speech-to-text generative performance. 
To generate accurate segmentation for the speech and timestamp, we adopt a purely data-driven multi-task learning approach. 
The training includes various tasks such as ASR, MT, transliteration, sentence segmentation and ITN, and translation interpretation. Motivated by Whisper, timestamp prediction is also integrated into our framework.
In addition, we apply the Chain-of-Thought~\cite{wei2022chain} approach for instruction tuning using the ``speech-transcription-translation'' dataset.
We trained our model on large datasets of English and Chinese audio and text, conducting experiments on English$\leftrightarrow$Chinese ST. 
Automatic evaluation results highlight the outstanding performance of our model. 
For long-form speech translation, human evaluation further confirms that our model surpasses the commercial cascaded systems, particularly excelling in cases that involve leveraging speech prosody, context, as well as translating code-switching and specialized terminology.
Our contributions are:
\begin{itemize}
    \item We propose \method, which leverages large language model in speech translation task. 
    It is capable of processing long audio inputs and producing accurately timestamped transcriptions and translations in a unified model.


    \item We conduct large-scale training and experiments on English-Chinese ST and find that \method can achieve generate better translation than the cascade system and industrial models.

\end{itemize}

\section{Method}
In this section, we describe how the model is constructed, including three aspects: model architecture~(Sec.~\ref{sec:method:frame}), training methods~(Sec.~\ref{sec:method:train}), and data preparation~(Sec.~\ref{sec:method:data}).

\begin{figure}[t]
    \centering
    \includegraphics[width=0.8\linewidth]{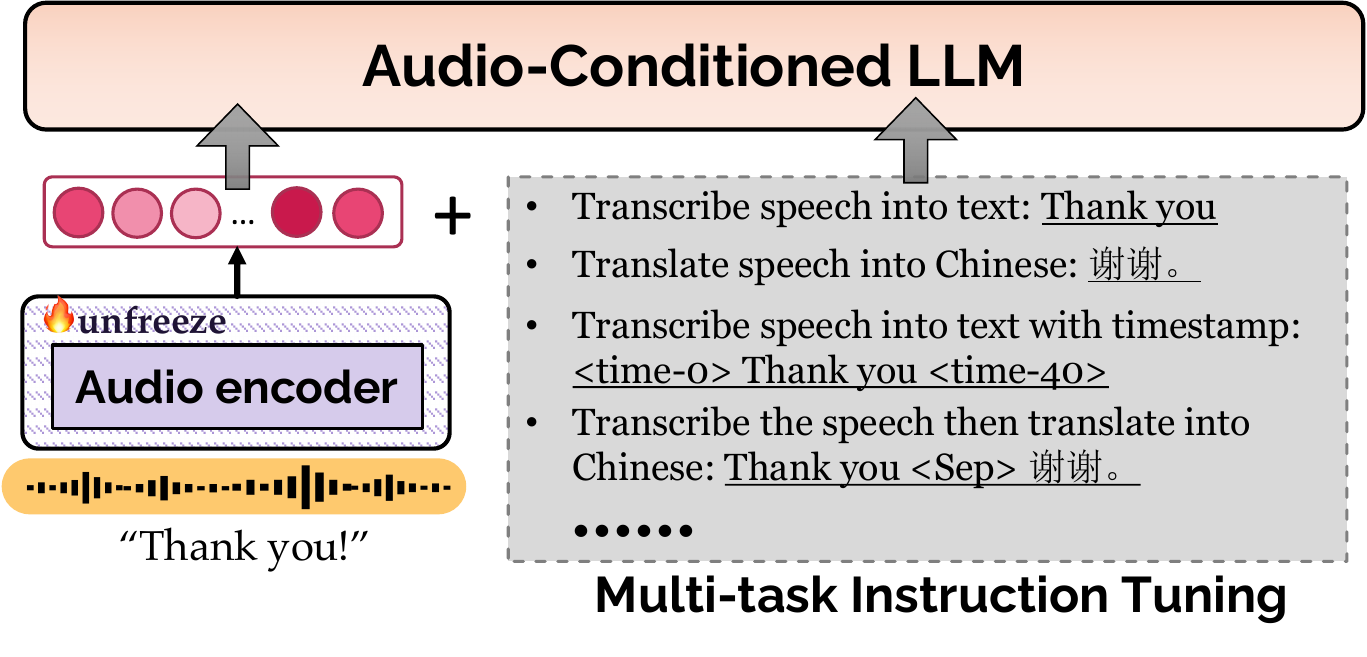}
    \caption{Framework of \method.}
    \label{fig:framework}
\end{figure}

\subsection{Framework}
\label{sec:method:frame}
As illustrated in Figure~\ref{fig:framework}, the model architecture comprises a Audio Encoder to convert audio input into representations, and an Audio-Conditioned LLM that accepts these representations, transforming the audio into textual output based on relevant instructions.

\noindent\textbf{Audio Encoder}~
We adopted the same structure as the Whisper Encoder~\cite{radford2023whisper}, which is based on a Transformer Encoder architecture: it begins with two convolutional layers with a stride of 2 and a kernel size of 3 to downsample the input audio, and the downsampled representation is combined with Sinusoidal position embeddings before being fed into the Transformer layers. For input, the audio signal is sampled at 16k Hz and converted into an 80-channel log Mel spectrogram. We used the same settings as \texttt{Whisper-Large-v2}\footnote{\url{https://huggingface.co/openai/whisper-large-v2}}, with 32 Transformer layers, a width of 1280, and 20 heads.
Unlike the Whisper model, we removed the restriction of padding input lengths to a constant 30 seconds, allowing for audio of any length. We utilized the parameters from the \texttt{Whisper-Large-V2} model for initialization, but during the tuning phase, we unfroze the parameters, allowing them to be optimized together.

\noindent\textbf{(Audio-conditioned) LLM}~
We adopted the same Transformer decoder-only structure as GPT~\cite{brown2020language}. 
Given instructions and audio representations from the Audio Encoder, it outputs the corresponding text response, expressed as~$p(y|\text{instruction},\text{AudioEnc}(x))$, where $x$ is the speech input. The instructions contain various related tasks, which will be introduced in Section~\ref{sec:method:train}.
To cater to the demands of larger data and industrial deployment, we utilized the model size with 13 billion parameters, whose width and depth are the same as the GPT-3 13B model.

\subsection{Data}
\label{sec:method:data}
We utilized extensive text and speech data to train the model.
We pre-train the LLM with about 450b tokens of Chinese and English raw text. 
During the tuning phase, our data consisted of pure text, text translation, speech translation, and speech recognition. 

\noindent\textbf{For text data}:~
We used about 130 million of Chinese-English translation pairs, encompassing both paragraph and sentence-level translations. 
We employed pyPinyin\footnote{\url{https://github.com/mozillazg/python-pinyin}} to convert Chinese to Pinyin and Espeak\footnote{\url{https://espeak.sourceforge.net}} for English to IPA, allowing us to form English $\leftrightarrow$ IPA $\leftrightarrow$ Pinyin $\leftrightarrow$ Chinese datasets. 
Beyond translations, we manually annotated thousands of sentences with translation notes.

\noindent\textbf{For speech recognition}:~
We utilized roughly 120k hours of English ASR data, 10k hours of Chinese data, and 5k hours of mixed Chinese-English data.

\noindent\textbf{For speech translation}:~
Recognizing the scarcity of well-annotated large-scale speech translation datasets, we employed GigaST~\cite{ye2023gigast} -- a 10k-hour dataset expanded from English ASR into a Pseudo ST dataset. Similarly, we obtained 10k hours of Chinese-to-English ST data expanded from WeNetSpeech~\cite{zhang2022wenetspeech}. Additionally, we incorporated some genuine speech translation data, primarily from CoVoST-2~\cite{wang2021covost} and MuST-C~\cite{cattoni2021must}.

\subsection{Training Method}
\label{sec:method:train}

\noindent\textbf{Pretraining}~
During pre-training, we first pre-trained the LLM using vast amounts of Chinese and English monolingual text corpora. 
For the Audio Encoder, we initialized with parameters from \texttt{Whisper-Large-V2}, which was weakly supervised pre-trained on 680k hours of audio and ASR transcriptions~\cite{radford2023whisper}.

\noindent\textbf{Instruction Tuning}~
We extensively perform instruction tuning across various tasks related to speech and text, where the instruction examples can be found in Table~\ref{tab:task_instruction}.
\begin{itemize}[itemsep=0em, parsep=0em, leftmargin=2em]
    \item \textbf{Machine Translation (MT)}: Specifically referring to text translation, where the input is the source language text and the output is the translated text in the target language. 
    
    \item \textbf{Automatic Speech Recognition (ASR)}: This encompasses both Chinese and English speech recognition. Transcriptions are typically devoid of punctuation and case distinctions, and the text should precisely match the audio content, including filler words like "uh" or "ummm". 
    
    \item \textbf{Speech Translation (ST)}: This includes bidirectional speech translation between Chinese and English. ST datasets not only have the translated target language but also the source language transcription. Given the scarcity of high-quality, well-aligned real-world speech translation data, we also include pseudo-ST data for training. These target language translations are derived from the source language transcription via commercial machine translation APIs. To distinguish, we employ the term ``naturally'' in the instruction. As we possess "speech-transcription-translation" supervised data, we can use a step-by-step transcription-then-translation method for instruction tuning, further detailed in the "\textbf{Chain-of-Thought Instruction}" section.

    \item \textbf{Pronunciation Translation}: Apart from text and speech, we've also observed an intermediate form: pronunciation. For English, we convert text to International Phonetic Alphabet (IPA) symbols, and for Chinese, we turn text into Chinese Pinyin. As shown in Table~\ref{tab:task_instruction}, the instructions regarding the mutual conversion between speech, pronunciation, and text can be either text-to-phoneme instructions or audio-to-phoneme instructions. And we even employ more complex combinations, generated step-by-step through a Chain-of-Thought Instruction.

    \item \textbf{Inverse Text Normalization (ITN) and Smoothing}: Recognizing that many ASR datasets have transcriptions without punctuation and case distinction, transforming the transcribed text into a more readable format aids in aligning with translations, thereby enhancing translation quality. As such, we introduced the ITN and smoothing tasks to make transcribed texts more reader-friendly. 
    
    \item \textbf{Translation Explanation}: We find that explanations for translations, especially those with specialized terminology, can offer valuable insights and enhance translation accuracy. These translation explanations are added as supplementary annotations after the translation. Specifically, we selected about thousands of sentences containing proper nouns or specialized knowledge in the ST data for annotation and incorporated them into the training set.

    \item \textbf{Timestamped Speech-to-Text Generation}: To endow the model with segmentation and timestamp prediction capabilities, we incorporated timestamped generation tasks. 
    We add timestamp tokens at the beginning and end of each transcription (or translation) training sample. Similar to Whisper's data processing~\cite{radford2023whisper}, we quantized timestamps to the nearest 20 milliseconds.

\end{itemize}

\begin{table}[]
    \centering
    \begin{tabular}{p{2.5cm}|p{11cm}}
        \toprule
        \textbf{Tasks} & \textbf{Instruction Examples} \\
        \midrule
        \multicolumn{2}{l}{\textbf{Standard Instructions}} \\
        \midrule
        MT & \{Text\} Translate the text into Chinese:\{translation\}.\\ \midrule
        ASR &  \texttt{[SpeechRepr]} Transcribe the speech into English:\{transcription\}. \\ \midrule
        \multirow{2}{*}{ST} & \texttt{[SpeechRepr]} Translate speech into English:\{translation\}.\\
        & \texttt{[SpeechRepr]} Translate speech into English naturally:\{translation\}. \\
        \midrule
        Pronunciation  & \{Text\} Translate the text to English IPA: \{IPA\}\\
        Translation & \texttt{[SpeechRepr]} Transcribe the speech into Chinese Pinyin: \{Pinyin\}. \\
        \midrule
        ITN \& Smoothing & \texttt{[SpeechRepr]} Transcribe the speech to standard English: \{smoothed\_transcription\}. \\
        \midrule
        Translation Explanation & \texttt{[SpeechRepr]} Translate the speech into English and explain it: \{translation\} \{explanation\}. \\
        \midrule
        Timestamped ASR & \texttt{[SpeechRepr]} Transcribe the speech to Chinese text with timestamp: \{transcription\_with\_time\}.\\
        \midrule
        Timestamped ST & \texttt{[SpeechRepr]} Translate the speech to English text with timestamp: \{translation\_with\_time\}.\\
        \midrule
        \multicolumn{2}{l}{\textbf{Chain-of-Thought Instructions}} \\
        \midrule
        ST & \texttt{[SpeechRepr]} Transcribe speech and translate it into English: \{transcription\} \{translation\}.\\ 
        \midrule
        ITN \& Smoothing & \texttt{[SpeechRepr]} Transcribe the speech and then perform inverse text normalization: \{transcription\} \{smoothed\_transcription\}. \\
        \midrule
        Translation Explanation  & \texttt{[SpeechRepr]} Transcribe speech, translate it into English and explain the translation: \{transcription\} \{translation\} \{explanation\}. \\
        \midrule
        Timestamped ST & \texttt{[SpeechRepr]} Transcribe the speech and then translate it to English text with timestamp: \{transcription\_with\_time\} \{translation\_with\_time\}.\\

        \bottomrule
    \end{tabular}
    \caption{List of the instruction examples for different tasks. \texttt{\texttt{[Text]}} represents the source text and \texttt{[SpeechRepr]} denotes the speech representation. }
    \label{tab:task_instruction}
\end{table}

\noindent\textbf{Chain-of-Thought Instruction}~
Previous studies have suggested that allowing LM to think step by step can enhance their performance in complex tasks such as mathematical problem-solving and reasoning~\cite{kojima2022large,wei2022chain}, termed as Chain-of-Thought~(CoT)~\cite{wei2022chain}. 
Human interpreters typically first listen and understand what the speaker is saying before translating. 
For ST tasks, we can decompose the task into first performing ASR to recognize the text and then providing the translation. 
This technique of recognizing first and then translating has been utilized in many speech-to-text cross-modal large models~\cite{zhang2023speechgpt}. 
We further refine this approach by not only transcribing to natural language but also adding: 
\begin{inparaenum}[\it 1)]
    \item transcription to IPA (or Pinyin) followed by translation; 
    \item translation (of text or speech) followed by explanation; 
    \item or even longer chain, \textit{e.g.} transcription first, followed by ITN smoothing, then translation, and finally explanation.
\end{inparaenum}

\noindent\textbf{In-Context Training}~
In machine translation, context information is crucial, especially in certain situations such as handling ambiguous words and maintaining sentence coherence~\cite{carbonell2006context,zheng2020toward,sun2022rethinking}.
This is also true for speech translation~\cite{zhang2021beyond}. 
However, including all historical speech segments, transcriptions, and translations in training can be cumbersome and burdensome for the model training. 
Thus, we only append contextual \textbf{text} to the instruction. 
The training data mainly comes from long speeches and videos. We select the preceding $N$ sentences of the current speech clip for training.

\subsection{Inference for Long Speech Translation}
\label{sec:method:long_inference}
Our model is adept at translating long audio files of indefinite length in a complete end-to-end manner from speech to text. 
The generation process unfolds through a series of steps:
\begin{enumerate}[itemsep=0em, parsep=0em, leftmargin=2em]
    \item It begins by extracting a 30-second slice from the beginning of the audio input. This time-bound segment serves as a manageable unit for the model to process without compromising on the context of the ongoing speech.

    \item The model then processes this audio slice to produce corresponding timestamps, ASR and ST results. The generation of timestamp is crucial as it anchors the transcribed and translated text to specific segments of the audio, ensuring synchronicity between the spoken words and the text output.

    \item After processing a slice, the model evaluates the results associated with the last time token, a special token for timestamp, indicating the end of the processed segment. The audio preceding this last time token is then discarded, and the model proceeds to extract the next 30-second audio slice. This new segment now becomes the focus of the subsequent round of processing.
\end{enumerate}

This process repeats in a loop, with each iteration discarding processed audio and moving on to a new slice, until the entire audio file has been fully handled. By continuously updating the starting point based on the last time token, the model seamlessly stitches together a coherent and comprehensive translation of the entire audio stream, irrespective of its length. 
This inference procedure not only allows for the efficient handling of long audio but also ensures that the translation process is dynamic and adaptive, adjusting to the nuances of the speech as it progresses.

\section{Experiment}

\subsection{Experimental Setups}
\begin{tcolorbox}[width=1\textwidth,title=The prompt used to build the Cascade ST model with ChatGPT]
Instruction: I need you to perform the task of speech translation, and now you are given the recognized text corresponding to the speech, i.e. ASR result. The ASR result may be without punctuation or may be recognized as homophone, and then polish the ASR results, and try to be fluent. If you find some weird, unnatural sentences, that would be the errors from ASR. Try to fix them and guess the original sentences. Then translate it into \textbf{\{target\_language\}}, making sure that the fewer errors the better. \\
\\
The ASR result is\\
\texttt{```}\\
\textbf{\{ASR\_result\}}\\
\texttt{```}\\
\\
Please give me the polished ASR result and the \textbf{\{target\_language\}} translation that meets the above requirements, using the JSON format with the following keys: \\
- polished\_ASR\_result\\
- translation
\end{tcolorbox}

\noindent\textbf{Baseline ST systems.}~
We compared our model with several strong baselines, including end-to-end speech translation and cascaded speech translation approaches.
The end-to-end speech translation models include: 
\begin{inparaenum}[\it 1)]
    \item Whisper for direct speech translation\footnote{We use \texttt{Whisper large-v2} as in~\url{https://github.com/openai/whisper/discussions/661}, and ditto.},
    \item SeamlessM4T~\cite{barrault2023seamlessm4t}, 
    \item AudioPaLM~\cite{rubenstein2023audiopalm}, 
    \item XSTNet~\cite{ye2021end} model with \texttt{Wav2vec2.0-large} as the speech encoder and transformer-xl as the translation module (total parameter size = 800M) trained based on a large amount of MT, ST, and ASR.
\end{inparaenum}
The cascaded models include: 
\begin{inparaenum}[\it 1)]
    \item Whisper for ASR cascading with Google translation,
    \item Whisper for ASR cascading with ChatGPT for translation,
    \item A commercial cascaded system that is specially optimized for video translation, noted as ``ST Product'', whose ASR model is noted as ``ASR Product''.
\end{inparaenum}

\noindent\textbf{DataSets.}~ 
We conducted experiments on various datasets, encompassing public benchmark and in-house datasets.
\begin{itemize}[itemsep=0em, parsep=0em, leftmargin=2em]
    \item \textbf{For English-to-Chinese ST}: The public test sets primarily consisted of GigaST~\cite{ye2023gigast} and MuST-C v2~\cite{cattoni2021must}. GigaST includes podcasts and YouTube domains, while MuST-C is mainly derived from TED talks. 
The in-house test sets spanned three domains: \texttt{news}, conference \texttt{talks}, and \texttt{videos}. Each domain has distinct characteristics: the news domain features formal English, conference talks contain numerous technical terms, and videos, are more casual and recorded in noisier environments. 
We also conducted tests for long audio speech translation for En-Zh direction, primarily focusing on in-house test sets, \texttt{Long Talk} and 
\texttt{Long Video}. There are 46 and 35 audio segments in each, with an average duration of 2 minutes. The long audio speech translation tests were primarily compared with the commercial cascaded system, ST product.

    \item \textbf{For Chinese-to-English ST}: The public test set is the Zh-En subset of CoVoST2~\cite{wang2021covost}, sourced from CommonVoice recordings. Though the recordings are clear, many contain classical Chinese texts, making content comprehension challenging. 
Our in-house datasets also covered three domains: news, conference talks, and live-streamed videos. The news domain uses formal language, conference talks have many technical terms, and live-stream videos feature background noise and trending terms.
\end{itemize}

\noindent\textbf{Metrics.} 
For the quality of speech translation, we employ automated measurements using the BLEU and COMET~\cite{rei2020comet} scores. BLEU is an n-gram-based metric. However, existing research suggests that BLEU often does not reflect the alignment between translations and human expectations well~\cite{kocmi2021ship,yuan2021bartscore,sai2022survey}. Therefore, we recommend placing more emphasis on the COMET score.
In addition, we also perform the human evaluation for passage-level long-audio speech translation.
For automatic speech recognition, we evaluate using word error rate (WER) and for Chinese, we use character error rate (CER).

\begin{table}[t]
    \centering
    \footnotesize
    \begin{tabular}{l|cc|ccc|c}
        \toprule
        & \multicolumn{2}{c}{\texttt{\textbf{Public}}} & \multicolumn{3}{|c|}{\texttt{\textbf{In-house}}} & \multirow{2}{*}{Avg}\\
        \cmidrule{2-3} \cmidrule{4-6}
        Models & \texttt{GigaST} & \texttt{MuSTC-v2} & \texttt{News} & \texttt{Talk} & \texttt{Video}  &    \\
        \midrule
        \textit{End-to-end Models} \\
        SeamlessM4T & 26.4 / 78.0 &	21.1 / 79.6 & 26.4 / 82.3 & 25.0 / 81.0 & 23.0 / 75.1 & 24.3 / 79.2\\ 
        XSTNet~(800M) & 40.5 / 83.9 & 27.3 / 84.1 & 36.8 / 86.4 & 30.3 / 80.2 & 36.2 / 77.6 & 34.2 / 82.4 \\
        \midrule
        \textit{Cascaded Systems} \\
        Whisper + Google Trans & 38.8 / 84.2 & 26.7 / 83.7 & 36.1 / 86.0 & 33.9 / 86.1 & 35.6 / 84.8 & 34.2 / 85.0\\
        Whisper + GPT3.5 & 32.7 / 84.8 & 25.2 / 84.6  & 32.3 / 86.6 & 31.4 / 86.5 & 32.7 / 84.6 & 30.9 /  85.4\\
        ST product & \textbf{40.9} /  82.4 &	28.2 /  83.6 & 36.4 / 84.2 & 31.1 /  84.6 & 	37.5 /  84.1 & 34.8 / 83.8 \\
        \midrule
        \method & 39.6 / \textbf{85.3} & \textbf{29.6} / \textbf{86.0} & \textbf{36.6} / \textbf{87.2} & \textbf{36.2} / \textbf{86.8} & \textbf{39.8} / \textbf{85.7} & \textbf{36.4} / \textbf{86.2} \\
        \quad - CoT & 38.7 / 85.1 & 29.3  / 85.7 & 35.4 / 86.8 & 35.9 / \textbf{86.8} & 32.9 / 84.6 & 34.4 / 85.8 \\
        \quad ASR + MT & 39.5 / 83.9 & 28.8 / 85.5 & 36.0  / 86.2 & 35.1 / 86.1 & 38.8  / 84.8 & 35.6 / 85.3 \\
        \bottomrule
    \end{tabular}
    \caption{BLEU and COMET scores of different systems for En-Zh translation}
    \label{tab:st_enzh_bc}
\end{table}

\begin{table}[t]
    \centering
    \footnotesize
    \setlength\tabcolsep{3.0pt}
    \begin{tabular}{l|c|ccc|c}
        \toprule
        & \multicolumn{1}{c}{\textbf{\texttt{Public}}} & \multicolumn{3}{|c|}{\textbf{\texttt{In-house}}} & \multirow{2}{*}{Avg}\\
        \cmidrule{2-2} \cmidrule{3-5}
        Models & \texttt{CoVoST2} &  \texttt{News} & \texttt{Talk} & \texttt{Live}  &    \\
        \midrule
        \textit{End-to-end Models} \\
        SeamlessM4T & 22.3 / 77.1  & 22.5 / 80.5 & 18.4 / 79.8 & 13.3 / 72.1 & 19.5 / 77.5 \\ 
        Whisper ST & 17.0 / 69.7 & 25.1 / 79.6 & 20.5 / 79.4 & 17.5 / 72.6 & 20.0 / 75.3\\
        SpeechLLaMA & 12.3 / -  & - & - & - & - \\
        AudioPaLM-2 (8B) & 25.5 / - & - & - & - & - \\
        mSLAM-CTC (2B) & 10.0 / - & - & - & - & - \\
        \midrule
        \textit{Cascaded Systems} \\
        Whisper + Google Trans & 26.7 / 79.2  &  42.1 / 86.1 & 24.4 / 83.7 & 25.3 / 78.6 & 29.6 / 81.9 \\
        Whisper + GPT3.5 & 21.6 / 79.0 & 33.3 / 86.3 & 20.5 / 84.5 & 19.5 / 78.9 & 23.7 / 82.2 \\
        ST product & 29.9 / 80.5 & \textbf{45.9} / 86.5 & 23.2 / 82.5 & 27.7 / 80.9 & 31.7 / 82.6\\
        \midrule
        \method & \textbf{37.6} / \textbf{83.7} & 44.9 / \textbf{87.2} & \textbf{25.9} / \textbf{84.6} & \textbf{28.2} / \textbf{81.4} & \textbf{34.1} / \textbf{84.2}\\
            \quad - CoT & 17.8 / 82.7 & 42.9 / 86.7 & 25.5 / 84.3  & 26.6 / 80.1 & 28.2 / 83.5\\
            \quad ASR + MT & 15.5 / 82.1 & 35.4 / 85.4 & 20.9 / 83.5  & 19.9 / 78.3 & 22.9 / 82.3 \\   
        \bottomrule
    \end{tabular}
    \caption{BLEU and COMET scores of different systems for Zh-En translation}
    \label{tab:st_zhen_bc}
\end{table}

\begin{table}[t]
    \centering
    \footnotesize
    \setlength\tabcolsep{3.0pt}
    \begin{tabular}{l|ccc|cccc}
        \toprule
        & \multicolumn{3}{c}{\textbf{\texttt{Public}}} & \multicolumn{3}{|c}{\textbf{\texttt{In-house}}} & \multirow{2}{*}{Avg}\\
        \cmidrule{2-4} \cmidrule{5-7}
        Models & \texttt{GigaST} & \texttt{MuSTC-v2} & \texttt{LibriSpeech} & \texttt{News} & \texttt{Speech} & \texttt{Video}  &    \\
        \midrule
        SeamlessM4T & 22.36 & 8.40 & 3.43 & 17.75 & 14.34 & 27.77 & 15.68 \\ 
        Whisper & 13.39 & 7.67 & 2.70 & 12.74 & 9.69 & 15.03 & 10.20 \\
        ASR Product & 10.58 & \textbf{5.21} & 3.49 & \textbf{9.98} & \textbf{8.39} & \textbf{11.08} & 8.15 \\
        \midrule
        \method & \textbf{9.16} & 5.24 & \textbf{2.01} & 11.22 & 8.55 & 11.15 & \textbf{7.89} \\
        \bottomrule
    \end{tabular}
    \caption{WER of English ASR}
    \label{tab:asr_en}
\end{table}
\begin{table}[t]
    \centering
    \footnotesize
    \setlength\tabcolsep{3.0pt}
    \begin{tabular}{l|cc|cccc}
        \toprule
        & \multicolumn{2}{c}{\textbf{\texttt{Public}}} & \multicolumn{3}{|c}{\textbf{\texttt{In-house}}} & \multirow{2}{*}{Avg}\\
        \cmidrule{2-3} \cmidrule{4-6}
        Models & \texttt{CoVoST2} & \texttt{CommonVoice} &  \texttt{News} & \texttt{Speech} & \texttt{Live}  &    \\
        \midrule
        SeamlessM4T & 14.01 & 14.73 & 13.71 & 13.79 & 20.39 & 15.33 \\ 
        Whisper & 12.98 & 14.77 & 8.94 & 11.71 & 17.10 & 13.10\\
        ASR Product & 10.42 & 11.99 & \textbf{8.04} & \textbf{12.28} & 12.80 & 11.10 \\
        \midrule
        \method & \textbf{8.49} & \textbf{7.90} & 12.40 & 13.10 & \textbf{10.31} & \textbf{10.44} \\
        \bottomrule
    \end{tabular}
    \caption{CER of Chinese ASR}
    \label{tab:asr_zh}
\end{table}

\subsection{Main result}
\noindent\textbf{Speech Translation}~
Tables~\ref{tab:st_enzh_bc} and~\ref{tab:st_zhen_bc} list the results of the automatic evaluation metrics (BLEU and COMET) for different models in English-to-Chinese and Chinese-to-English speech translation, respectively. 
Our method achieved commendable performance, even surpassing commercial ST models and the strong Whisper+ChatGPT cascade system. 
As an end-to-end model, our model also performed better than open-source end-to-end models (or the models with reported BLEUs on CoVoST2 Zh-En subset).
In Table~\ref{tab:st_enzh_bc}, we compared the Encoder-Decoder~(XSTNet) and decoder-only \method using an equal magnitude of training data. The \method performed better in translation, due to the influence of model size as well as the improved fluency brought about by pretraining on large-scale corpora.
Of course, as a technical report of industrial practice, we acknowledge that we did not conduct a completely fair comparison (considering resource conservation, we did not undertake an exhaustive and fair comparison), due to differences in the data and model parameters used.
For instance, models like SeamlessM4T, Whisper, and AudioPaLM-2 all focus on multilingual training, involving respectively 470k, 680k, and 18k hours of multilingual speech in their training, while we focused on optimizing for English-Chinese bilingual translation, with approximately 52k hours of speech trained. Nevertheless, it is still noteworthy that for the Chinese-English speech-to-text data used in training, SeamlessM4T involves 18k hours of Chinese-English ST data and Whisper involves 11k hours, both not less than our model.


\begin{wraptable}{r}{0.5\linewidth}
    \centering
    \footnotesize
    \begin{tabular}{lcc}
         \toprule 
         & \texttt{Long Talk} & \texttt{Long Video}  \\ 
         \midrule
         ST product & \textbf{33.5} /  84.2  & \textbf{41.0} / 81.8\\
         Whisper + GPT3.5 & 28.5 / \textbf{86.3}  & 34.9 / 82.0 \\
         \midrule
         \method & 33.1 / \textbf{86.3} & 37.6 / \textbf{82.8} \\ 
         \quad w/o context & 32.6 / 85.2 & 37.5 / 80.7 \\
         \bottomrule
    \end{tabular}
    \caption{BLEU and COMET scores of the English-to-Chinese long Speech translation corpus.}
    \label{tab:long_st_context}
    
\end{wraptable}

\begin{table}[]
    \centering
    \footnotesize
    \begin{tabular}{l|ccc|ccc}
        \toprule
         & \multicolumn{3}{c}{\texttt{Long Talk}} & \multicolumn{3}{|c}{\texttt{Long Video}}\\
        \cmidrule{2-4} \cmidrule{5-7}
        & Precision & Recall & F1 & Precision & Recall & F1 \\
        \midrule

         Whisper & 92.56\% & 98.35\% & 95.36\% & 69.09\% & 97.09\%  & 80.71\% \\
         \method & 97.78\% & 94.61\% & 96.17\% & 83.66\% & 89.59\% & 86.52\% \\
        \bottomrule
    \end{tabular}
    \caption{Time Stamp Accuracy}
    \label{tab:time_stamp}
\end{table}

\noindent\textbf{Automatic Speech Recognition}~ 
As our multi-task training involved the ASR task, we also presented the performances for Chinese and English speech recognition in Tables~\ref{tab:asr_en} and~\ref{tab:asr_zh} respectively. 
Our model achieved lower WER and CER, and on average, even made fewer errors than commercial ASR models across various test sets. 


\begin{table}[]
    \centering
    \footnotesize
    \begin{tabular}{l|cc}
        \toprule
         & \texttt{Long Talk} & \texttt{Long Video} \\
        \midrule
         ST product & 3.84 & 3.87 \\
         \method & 4.06 & 4.02 \\
        \bottomrule
    \end{tabular}
    \caption{Human evaluation}
    \label{tab:human_eval}
\end{table}

\noindent\textbf{Speech Translation for long audio}~ 
Both the automatic evaluation results in Table~\ref{tab:long_st_context} and the huamn evaluation results in Table~\ref{tab:human_eval} indicate that \method performs better than cascaded systems in long-form speech translation tasks. 
This not only attests to the model's semantic translation capabilities but also indirectly validates the effectiveness of \method in timestamp prediction. 
More importantly, our model demonstrates a more concise and efficient performance in inference and deployment compared to cascaded systems. 

\noindent\textbf{Timestamp for Long Audio.} In order to verify the accuracy of our timestamps, we compare the VAD information predicted by \method to the groundtruth VAD label and calculate its Precision, Recall and F1 in Table~\ref{tab:time_stamp}. As the groundtruth VAD information is hard to obtain for our test set, we use VAD label predicted by a mature commercial systems, \texttt{ST Product}, whose timestamps are widely used in real-world softwares, as the groundtruth label. As shown in Table~\ref{tab:time_stamp}, \method is more accurate the timestamps predicted by Whisper.


\subsection{Ablations}

Given resource constraints, we didn't conduct ablation on every detail but focused on four key components: the introduction of extra tasks associated with speech translation, the CoT instruction tuning, the introduction of context during training, and the influence of real high-quality ST data.


\noindent\textbf{Chain-of-Thought Instruction Tuning improves Performance}
In both Chinese-to-English~(Table~\ref{tab:st_enzh_bc}) and English-to-Chinese~(Table~\ref{tab:st_zhen_bc}) experiments, 
we employed a stepwise Chain-of-Thought inference that involved speech recognition followed by translation. Compared to directly prompting the model to output speech translation, the CoT inference indeed yields better results.
During the generation of translations, the model benefits from useful context derived from the predicted transcriptions, thereby enhancing the accuracy of the translations.
Moreover, given that our model is capable of performing multiple tasks, we can also first conduct ASR to obtain transcriptions and then utilize the model for text translation, simulating the approach of a cascade system. 
Trained on the same large-scale dataset, an interesting finding is that CoT inference outperforms the cascaded inference. 
This improvement stems from the direct utilization of audio information. The intonation and prosody inherent in speech are more readily captured by an end-to-end model, which in turn positively impacts the quality of translation.
In Section~\ref{sec:exp:case}, we will use more practical cases to demonstrate the impact of audio information on translation intuitively.

\noindent\textbf{Context text matters}~
In Table~\ref{tab:long_st_context}, when removing the preceding predicted text as the context, we find that the performance drops a little bit, which proves that incorporating the context enhances translation, aligning with findings in the fields of machine translation and speech translation.



\subsection{Case Analysis}
\label{sec:exp:case}

\begin{CJK*}{UTF8}{gbsn}

\begin{table*}[th]
    \setlength{\belowcaptionskip}{-0.4cm}
    \centering
    \small
    \begin{tabular}{l|p{12cm}}
        \toprule
        \multicolumn{2}{c}{\textbf{CASE 1} } \\
        \midrule
        Transcription & you guys like dumplings? (in a rising intonation, indicating a question)\\
        ST product & 你们喜欢饺子\textcolor{red}{\underline{。}}\\
        \method & 你们喜欢饺子吗？ \\
        \midrule

        \multicolumn{2}{c}{\textbf{CASE 2} } \\
        \midrule
        Transcription & help help ahhhhh (in an alarmed tone)\\
        ST product & \textcolor{red}{\underline{帮助帮助。}}啊。\\
        \method & 救命！救命！ \\
        \midrule

        \multicolumn{2}{c}{\textbf{CASE 3} } \\
        \midrule
        Transcription & When it's tough, will you give up, or will you be relentless? \\
        ST product & 艰难的时候，你会放弃还是会\textcolor{red}{\underline{毫不留情}}？\\
        \method & 困难的时候，你会放弃，还是会坚持不懈？\\
        \midrule

        \multicolumn{2}{c}{\textbf{CASE 4} } \\
        \midrule
        Transcription & yeah i looked up before that, as long as you get the official taxi stand, then it should be about 100 rmb to the center of Beijing, \\
        ST product & 是的，在那之前我\textcolor{red}{\underline{抬头看了看}}。 只要你有官方的出租车站，那么到北京市中心应该大约需要100元人民币。 \\
        \method & 是的，我之前查过，只要你到官方的出租车停靠站， 到北京市中心大约需要100元人民币。  \\
        \midrule

        \multicolumn{2}{c}{\textbf{CASE 5} } \\
        \midrule
        Transcription & Fu Qi Fei Pian literally means slices of husband and wife's lungs. Eh. \\
        ST product & \textcolor{red}{\underline{Uchife pan}}字面意思是丈夫和妻子的肺部切片。 \\
        \method & 夫妻肺片，字面意思是丈夫和妻子的肺片。是的。\\
        \midrule

        \multicolumn{2}{c}{\textbf{CASE 6} } \\
        \midrule
        Transcription & It got this name because when the dish first originated, it used "Fei" or unwanted meats from cows, like cow tongues and lungs, and the dish was made best by a husband and wife duo. \\
        ST product & 它之所以有这个名字，是因为当这道菜最初起源时，\textcolor{red}{\underline{一个用过的Fey吃了奶牛不想要}}\textcolor{red}{的肉}，比如\textcolor{red}{\underline{凯尔顿}}和肺，这道菜是由一对夫妻做的最好的。 \\
        \method &  它之所以有这个名字，是因为当这道菜刚出现的时候，它使用了“废”的牛肉，比如牛舌和牛肺，这道菜是由一对夫妻共同制作的。 \\
        \midrule
        
        
        \multicolumn{2}{c}{\textbf{CASE 7} } \\
        \midrule
        Transcription & After bursting onto the fashion scene with the brand collective Vetements, Demna took the helm of Balenciaga in 2015 as part of a radical modernization, including the Triple S, which set a fire the ugly sneaker trend, and helped Balenciaga surpass a billion dollars in revenue in 2019, spawning a thousand copycats. \\
        ST product & 在凭借品牌闯入时尚界后\textcolor{red}{\underline{集体兽医购物中心}}，Demna于2015年掌舵巴黎世家，作为激进现代化的一部分， 包括Triple s， 它点燃了丑陋的运动鞋潮流，并帮助巴黎世家在2019年的收入超过10亿美元，催生了1000名模仿者。 \\
        \method &  在以品牌集合 Vetements 闯入时尚界后，德姆纳于 2015 年执掌巴黎世家， 作为激进现代化的一部分， 包括 Triple S， 它点燃了丑陋运动鞋的潮流， 并帮助巴黎世家在 2019 年的收入超过了 10 亿美元， 催生出 1000 个模仿者。 \\

        \bottomrule
    \end{tabular}
    \caption{Cases of English-to-Chinese ST. 
    The \textcolor{red}{\underline{red underlined text}} indicates incorrect or inappropriate translations.
    Case~1 and 2: \method enhances ST with speech prosody. Case~3 and~4: It disambiguates translations via context. Case~5-7: It excels in translating code-switch and entity-rich speech.
    }
    \label{tab:case_study}
\end{table*}

\end{CJK*}

\textbf{Prosody helps the translation.}~
\begin{CJK*}{UTF8}{gbsn}
In case 1, although the original input is in the form of a declarative sentence, the intonation in the audio is rising, indicating a questioning tone. 
\method provides the correct translation, while the commercial cascaded model ST product translates it into a declarative sentence due to the transcription output lacking a question mark. 
In case 2, the audio corresponding to "help help" is uttered in a shouting tone.
The cascaded model fails to discern the tone and translates it literally as ``帮助帮助。''. 
Our model, getting emotional information directly from the audio input, translates as ``救命！救命！'', which is sufficient to convey the speaker's terrified call for help.
\end{CJK*}

\textbf{Context helps to resolve ambiguity.}~
\begin{CJK*}{UTF8}{gbsn}
In Case 3, the term ``relentless'' can imply ``merciless'' (e.g., a merciless opponent) or ``unyielding'' (e.g., relentless pursuit). 
In the traditional cascade model, the MT model opts for a literal translation, rendering ``relentless'' as ``毫不留情'', but \method, aided by a more powerful language model, infers from the preceding phrase ``give up'' that ``relentless'' is an antonym, thus translating it as ``坚持不懈''.
In Case 4, while the literal translation of ``look up'' could indeed be ``抬头看了看'', the context suggests a more suitable translation is ``查过'' (means to check the information). Additionally, ``get the official taxi stand'' is used in conversational scenarios, and translating ``get'' as ``有''~(have) is inappropriate; it should be ``到达'' (to reach). 
For scenarios that need to take context into account, the \method model clearly does a better job.
\end{CJK*}

\textbf{Better translation for Code-switch speech.}~
\begin{CJK*}{UTF8}{gbsn}
Since our model is trained indiscriminately on both Chinese and English speech, it performs better in scenarios with mixed Chinese and English than commercial cascade models. 
For instance, in Cases 5 and 6, the speaker introduces a Sichuan dish called ``夫妻肺片'' (Chinese pinyin ``Fu qi fei pian''). This term is hard for an ASR model in English to recognize, whereas our model can translate it correctly. 
It is also a similar case for `Fei'' (the pronunciation of "废", meaning unwanted) in Case 6; the cascade model generates a translation completely ungrammatical and unreadable, 
Meanwhile, the \method's translation was very impressive. It does not translate literally but uses a homophonic and concise presentation: ``废''的牛肉. 
This capability partly stems from the joint training in multiple languages, and another important reason is the inclusion of the pronunciation translation task in the multi-task instruction tuning, with IPA and pinyin data greatly aiding the code-switch ST.
\end{CJK*}

\textbf{Better translation for name entities.}~
\begin{CJK*}{UTF8}{gbsn}
In Case 7, the translation encompassed a multitude of specialized terms from the fashion industry, such as brand names like Vetements and Balenciaga, as well as person name. First, we find that even the sophisticated commercial ASR models still struggle with recognizing name entities. 
For instance, the brand name Vetements was inaccurately recognized as the phonetically similar word ``veterinarian'', resulting in a completely erroneous translation of the brand collective as ``集体兽医购物中心''. 
In scenarios where it's unnecessary to identify specific proper nouns, such as ``cow tongues'' in Case 6, the ASR model mistakenly recognized as ``Kelton'', which led to the translation ``凯尔顿'', rendering the translation incomprehensible.
In contrast, our model leverages the power of pre-trained LLM and correctly translates the proper nouns when appropriate. 
This capability is likely enhanced by the inclusion of name entity explanation annotations in the translation explanation task of multi-task instruction tuning. 
These annotations assist the model in leveraging context and the extensive knowledge embedded in the pre-trained model to deliver more precise translations.
\end{CJK*}


\section{Conclusion}

In this paper, we introduced \method, a complete end-to-end ST system excelling in processing long audio sequences with precise timestamped outputs. 
Our model integrates LLM with speech representation from the Whisper encoder, and benefits from multi-task instruction fine-tuning, CoT instructions, and in-context training. 
\method significantly outperforms other open-source ST models and the cascade industrial systems in English$\leftrightarrow$Chinese ST tasks.

\section{Limitations and Broader Impacts}



\noindent\textbf{Limitations}~
We fully acknowledge that this paper has its limitations in experiments. 
For instance, while our method involved multiple tasks related to speech translation, due to resource constraints, we feel sorry that we couldn't validate the effectiveness of each task individually. 
Also, we couldn't perform fair comparisons ``with the same amount of training data or model size'' as typically required by the academic community, since this is a technical report rooted in industrial practice. 

\noindent\textbf{Broader Industrial Impacts}~
It has been seven years since the concept of E2E speech translation was first introduced. 
However, we've noticed that cascaded models remain the industry's best practice. 
Thus, we approached the idea of E2E speech translation from an industrial practice perspective.
We paid particular attention to comparing it with commercial cascaded models, including the Whisper ASR followed by the ChatGPT model. 
We hope this work provides some guidance for industrial practice and demonstrates to the community that deploying a complete E2E ST model in industry is indeed feasible. 
Additionally, we hope the idea of CoT inference can challenge the industrial adherence to the ``more controllable'' cascaded models --- proving that fully E2E models can also provide visibility into intermediate outputs.

\noindent\textbf{Future}~
For future research, we should consider various aspects, such as,
\begin{itemize}[itemsep=0em, parsep=0em, leftmargin=2em]
    \item \textbf{More efficient and effective tuning methods}: Under the 13b model scale, our model requires 32 A100 GPUs for instruction tuning. Although convergence requires only tens of thousands of steps, it takes about a week of training time. LLM pre-training requires even longer. Employing more effective tuning methods could significantly reduce iteration time, making ST research more accessible to the laboratory setting.

    \item \textbf{Multilingual speech translation}: While progress has been made in multilingual speech translation~\cite{barrault2023seamlessm4t,rubenstein2023audiopalm}, there's still a gap between industry application. Many practices in our method, such as pronunciation translation and translation explanation, are highly suitable for multilingual settings.

    \item \textbf{Speech-to-speech translation}: Progress has also been made in speech-to-speech translation within the LLM framework~\cite{zhang2023speak,dong2023polyvoice,wang2023viola}. A key question is whether it is worthwhile to discretize speech input. Discretized input may lose information that could be crucial for ST, such as prosody and intonation. In terms of speech generation, preserving voice timbre, controlling fine-grained emotions, and synchronizing lip movements in video scenarios are intriguing topics.

    \item \textbf{Streaming speech translation}: How to make the model understand where to provide correct semantic grouping with minimal delay to output high-quality translation is a challenge and not yet solved by LLMs alone.
\end{itemize}

\bibliographystyle{IEEEtran}
\bibliography{ref}


\end{document}